\documentclass[conference,final]{IEEEtran}
\usepackage{cite}
\usepackage{textcomp}
\usepackage{graphicx}
\usepackage{amsmath}
\usepackage{amssymb}
\usepackage{amsfonts}

\usepackage[english]{babel}
\usepackage{multirow}

\usepackage{subfigure}
\usepackage[font=small,labelfont=bf]{caption}
\usepackage{bm}
\usepackage{nicefrac}
\usepackage{microtype}

\usepackage{tikz}

\usetikzlibrary{
	arrows,
	arrows.meta,
	angles,
	babel,
	backgrounds,
	calc,
	decorations,
	decorations.pathreplacing,
	decorations.pathmorphing,
	fit,
	intersections,
	matrix,
	petri,
	positioning,
	shapes,
	spy,
	through,
	quotes,
}

\newcommand{\eg}{e.\,g.,\ }
\usepackage{cleveref}
\usepackage[absolute]{textpos}
\newcommand{\copyrightstatement}{
	\begin{textblock*}{17cm}(20mm,1mm)    
		\noindent
		\footnotesize
		\copyright 2019 IEEE. Personal use of this material is permitted. Permission from IEEE must be
		obtained for all other uses, in any current or future media, including
		reprinting/republishing this material for advertising or promotional purposes, creating new
		collective works, for resale or redistribution to servers or lists, or reuse of any copyrighted
		component of this work in other works.
	\end{textblock*}
}

\RequirePackage{tikz}
\tikzstyle{line} = [draw, -latex']
\tikzstyle{arrow} = [draw,-latex]

\def\P(#1){\Phelper#1|\relax\Pchoice(#1)}
\def\Phelper#1|#2\relax{\ifx\relax#2\relax\def\Pchoice{\Pone}\else\def\Pchoice{\Ptwo}\fi}
\def\Pone(#1){\Pr\left( #1 \right)}
\def\Ptwo(#1|#2){\Pr\left( #1 \mid #2 \right)}
\def\Ptwot[#1|#2]{\Pr\left( #1 \mid #2 \right)}
\def\Ptree(#1|#2,#3){\Pr\left( #1 \mid #2,#3 \right)}
\def\Pr{{p}}

\IEEEoverridecommandlockouts

\begin{document}
\copyrightstatement
\bstctlcite{bibcontrol_etal4}
\title{%
    Combining Deep Learning and Model-Based Methods for Robust Real-Time Semantic Landmark Detection
}


\author{%
    Benjamin Naujoks%
    \thanks{
		All authors are with the Institute for Autonomous Systems Technology (TAS) of the University of the Bundeswehr Munich, Neubiberg, Germany. Contact author email: benjamin.naujoks@unibw.de},
    Patrick Burger
    and %
    Hans-Joachim Wuensche
}

\maketitle

\begin{abstract}
    Compared to abstract features, significant objects, so-called landmarks, are a more natural means for vehicle localization and navigation, especially in challenging unstructured environments.
The major challenge is to recognize landmarks in various lighting conditions and changing environment (growing vegetation) while only having few training samples available.
We propose a new method which leverages Deep Learning as well as model-based methods to overcome the need of a large data set.
Using RGB images and light detection and ranging (LiDAR) point clouds,
our approach combines state-of-the-art classification results of Convolutional Neural Networks (CNN), with robust model-based methods by taking prior knowledge of previous time steps into account.
Evaluations on a challenging real-wold scenario, with trees and bushes as landmarks, show promising results over pure learning-based state-of-the-art 3D detectors, while being significant faster.

\end{abstract}
\section{Introduction}
Vehicle localization is an important ability for affordable autonomous vehicles in challenging environments to overcome the need of precise global positioning information. It typically derives from combinations of global navigation satelite systems (GNSS) and inertial navigation systems \cite{bena:skog2009car}.
Landmarks play an important role to utilize map-matching localization systems \cite{bena:levinson2007map,bena:yu2017vehicle}.
In these cases, a global position is estimated through the relative pose of the vehicle to detected landmarks, which are compared to corresponding landmarks in a digital map.
Another use case for landmarks are navigation systems for autonomous vehicles, especially in split convoy scenarios.
For instance, the leader vehicle communicates to the autonomously following vehicle to make a turn at a specific location, which can be described with landmarks.
Common landmarks are roads \cite{bena:hata2014,bena:hata2016,bena:vivacqua2018self} or signs \cite{bena:qu2015vehicle,bena:verentsov2017bayesian} as the typical scenarios are in urban areas.
We operate in unstructured environments with vegetation and without road markings and rare existence of buildings or signs.
Especially accurate maps are hardly available for those environments.
Therefore, we have to rely on trees and bushes as landmarks.
\begin{figure}
	\begin{center}
	\input{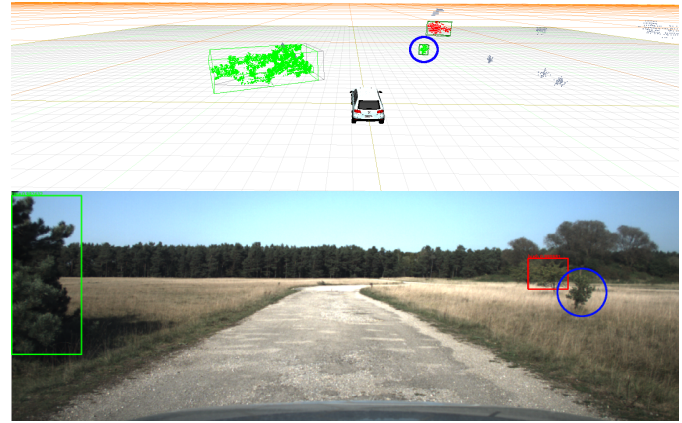}
	\caption{Result of our proposed architecture. The lower part shows detection results of the 2D CNN, whereas the upper part visualizes the \textbf{final result} with segmented and classified 3D landmarks. Here, classes are tree (green) and bush (red). Even though the little tree on the right side of lower part (marked with a blue ring) is not detected in the image of this time step, our approach robustly detects and classifies it through prior knowledge of previous time steps.}\label{fig:teaser}
	\end{center}
\end{figure}

Nowadays, the leading object detection and segmentation algorithms are Deep Learning based methods.
Well-known examples for the 2D case are the Single-Shot-Detector Network (SSD) \cite{bena:liu2016ssd}, Mask R-CNN \cite{bena:he2017mask} or YOLO9000 \cite{bena:redmon2017yolo9000}.
Surprisingly, the 3D case often is an overlooked task and only recently Convolutional Neural Networks for raw point clouds of depth sensors, \eg LiDAR, have been proposed \cite{bena:pointnet,bena:pointnet++}.
Although the PointNet architectures of \cite{bena:pointnet,bena:pointnet++} have proven to be reliable classifiers, they are not capable of instance-level segmentation.
\cite{bena:frustum} achieves instance-level segmentation and classification with frustum proposals in an end-to-end fashion.
Nevertheless, this approach tends to need a large training data set to robustly detect 3D landmarks, as they are highly dependent on continuous 2D object detections.
Instead, we propose to also consider model-based segmentation algorithms and recursive filtering.
Therefore, we are capable of handling fluctuating detection results of the learning-based 2D object detector by utilizing prior knowledge of previous time steps.

Our architecture outperforms the other methods on a challenging real-world scenario with few training samples available, while being real-time capable even on a consumer graphic card.


Summarizing, our main contributions are:
\begin{itemize}
	\item A novel architecture, utilizing RGB and depth information, to robustly detect and track semantic landmarks in challenging, unstructured environments.
	\item Outperforming results are shown for small training data sets.
	\item We quantitatively and qualitatively compare our method with state-of-the-art pure learning based methods to show the strengths and limitations of our approach.
	\item Real-time performance (below $ 100 $ms) even on a consumer graphic card like the NVIDIA GeForce GTX 1060.
\end{itemize}

\begin{figure*}[!t]
	\begin{center}
		\newsavebox\cnn
\begin{lrbox}{\cnn}
	\begin{tikzpicture}[]
	\begin{scope}[every node/.append style={yslant=-0.5},yslant=-0.5]
	\shade[right color=gray!10, left color=black!50] (0,0) rectangle +(1,1);
	\node at (0.5,0.5){\text{CNN}};
	\end{scope}
	\begin{scope}[every node/.append style={yslant=0.5},yslant=0.5]
	\shade[right color=gray!70,left color=gray!10] (1,-1) rectangle +(1,1);
	\draw (1,-1) -- (1,0);
	\draw (1.25,-1) -- (1.25,0);
	\draw (1.5,-1) -- (1.5,0);
	\draw (1.75,-1) -- (1.75,0);
	\draw (2,-1) -- (2,0);
	\end{scope}
	\begin{scope}[every node/.append style={yslant=0.5,xslant=-1},yslant=0.5,xslant=-1]
	\shade[bottom color=gray!10, top color=black!80] (2,1) rectangle +(-1,-1);
	\draw (2,0) -- (2,1);
	\draw (1.75,0) -- (1.75,1);
	\draw (1.5,0) -- (1.5,1);
	\draw (1.25,0) -- (1.25,1);
	\draw (1,0) -- (1,1);
	\end{scope}
	\end{tikzpicture}
\end{lrbox}

\begin{tikzpicture}
\tikzstyle{state}=[shape=circle,draw=red!50,fill=red!10,minimum size=1.5cm]
\tikzstyle{connection}=[inner sep=0,outer sep=0]
\tikzstyle{box}=[shape=rectangle,draw=green!60!black,fill=green!10!black,text width=1em]
\tikzstyle{smallbox}=[shape=rectangle,draw=green!60!black,fill=green!10!white]
\tikzstyle{lightedge}=[<-,dotted]
\tikzstyle{emptybox}=[shape=rectangle,draw=white,fill=white]
\tikzstyle{whitebox}=[shape=rectangle,draw=white,fill=white]
\tikzstyle{bigbox} = [draw=blue!50, thick, fill=blue!5, rounded corners, rectangle]
\tikzstyle{mainstate}=[state,thick]
\tikzstyle{mainedge}=[<-,thick]
\tikzset{
	main/.style={circle,minimum size=2cm, thick, draw =black!80, node distance = 8mm},
	connect/.style={-latex, thick},
	emptybox/.style={rectangle,minimum width =60pt, minimum height=40pt,fill=blue!5,node distance = 10mm},
	box/.style={rectangle, minimum width =60pt, minimum height=40pt,text width=5em,text centered, draw=gray!30!black,fill=gray!30!white,node distance = 10mm}
}

\node[state](imag){Camera};
\node[emptybox](ssd)[right=2.45cm of imag]{\usebox\cnn};

\path [line] (imag) -- node [text width=2.5cm,midway,above=0.1em,align=center ] {Image} (ssd);
\node[state](instances)[below=of imag]{LiDAR};
\node[smallbox](segmentation)[right=0.4cm of instances]{Clustering};

\node[connection](pcdump)[right=0.17cm of instances]{};
\node[connection](pcudump)[below=0.9cm of pcdump]{};
\draw[dotted] (pcudump) -- (pcdump);
\node[connection](segdump)[right=0.17cm of segmentation]{};
\node[connection](segudump)[below=0.9cm of segdump]{};
\draw[dotted] (segudump) -- (segdump);


\node[box](projection)[right=2.5cm of instances]{Projection};
\node[box,align=center](gdpf)[right=2.452cm of projection]{GDPF};
\path [line] (projection) -- node [text width=2.5cm,midway,above=0.1em,align=center ] {3D Instances with Class Proposals} (gdpf);
\path [line,align=center] (ssd) -- node [text width=5cm,midway,left=0.1em,align=right] {Classified 2D Object Proposals} (projection);
\path [line] (segmentation) -- node [text width=2.5cm,midway,below=2.5em,align=center ] {3D Instances} (projection);
\path [line] (instances) -- node [text width=2.5cm,midway,below=2.5em,align=center ] {Point Cloud} (segmentation);
\node[box](pointnet)[right=2.82cm of gdpf]{Point Cloud Classifier};
\path [line] (gdpf) -- node [text width=2.2cm,midway,above=0.1em,align=center ] {Segmented Point Clouds with Class Proposals} (pointnet);
\node[whitebox](dump)[below=of pointnet]{};
\node[whitebox](dump2)[below=of gdpf]{};
\draw [line] (pointnet) -- (dump.center) -- node [text width=2.3cm,midway,above=0.1em,align=center ] {Classified Components} (dump2.center) -- (gdpf);
\node[state,draw=red!10](input)[above=0.4cm of imag]{\textbf{Input}};
\node[emptybox,align=center](proposals)[above=0.1cm of ssd]{\textbf{Proposal} \\ \textbf{Generation}};
\node[emptybox](filter)[right=2.5cm of proposals]{\textbf{Tracking}};
\node[emptybox,align=center](pclass)[right=2.7cm of filter]{\textbf{Point Cloud} \\ \textbf{Classification}};
\begin{pgfonlayer}{background}
	\node[shape=rectangle,draw=red!50,thick,fill=red!10,rounded corners][fit = (input) (instances)]{};
\end{pgfonlayer}
\begin{pgfonlayer}{background}
	\node[bigbox] [fit = (proposals) (projection)] {};
\end{pgfonlayer}
\begin{pgfonlayer}{background}
	\node[bigbox] [fit = (filter) (gdpf)] {};
\end{pgfonlayer}
\begin{pgfonlayer}{background}
	\node[bigbox] [fit = (pclass) (pointnet)] {};
\end{pgfonlayer}
\end{tikzpicture}
		\caption{The general Semantic Landmark Detection Pipeline. The red circles denote the input sensors, which are RGB camera and LiDAR. The image is the input for the 2D object detection CNN, whereas data of the depth sensor is clustered to 3D bounding boxes by using the method of \cite{bena:burger_itsc2018}. In the projection step, the classified 2D objects are associated with the 3D instances of the clustering. Afterwards, the 3D instances are associated and tracked with a component of the GDPF \cite{bena:GDPF}. Then, the accumulated point cloud of every component is classified through a modified PointNet. Lastly, the GDPF gets the class of the component.}\label{fig:pipeline}
	\end{center}
\end{figure*}
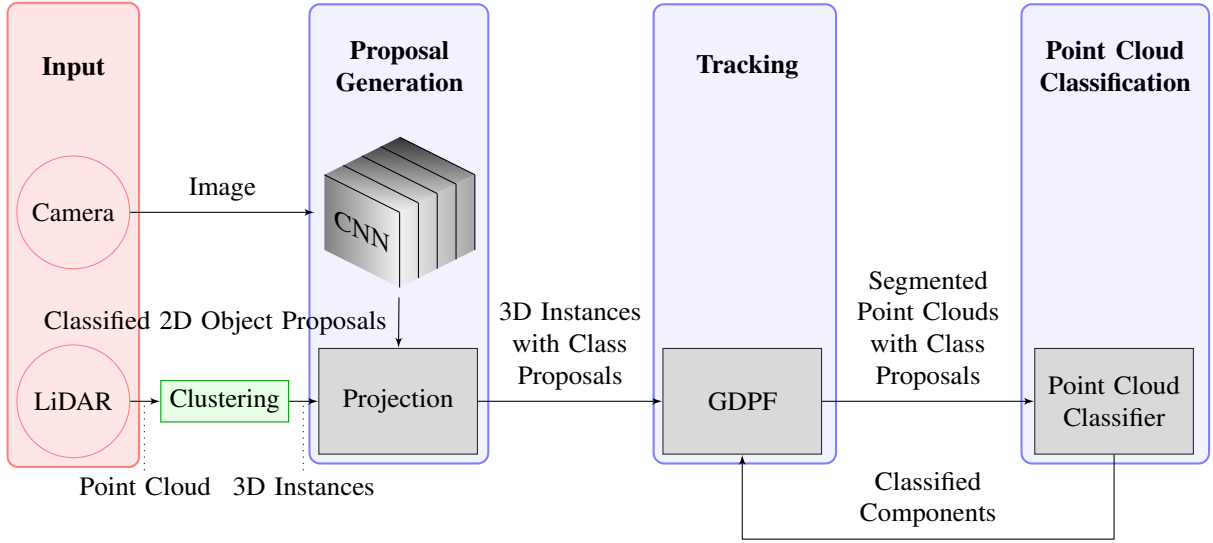
\section{Related Work} \label{sec:related_works}
\subsection{Landmark Detection for Vehicle Localization}
There have been two main types of landmarks in common research.
The first type are feature-based landmarks. Convolutional Neural Networks (CNN) have shown to be state-of-the-art feature generators. Many researchers have successfully used them in recent years \cite{bena:cramariuc2018learning,bena:elbaz20173d,bena:vincent2008extracting,bena:wohlhart2015learning}.
The second type are semantic landmarks.
They have the advantage of being robust against different point of views or small changes in the environments.
Typically semantic landmarks are signs \cite{bena:suhr2017sensor,bena:de2003traffic,bena:jiang1998robust}, pole-like structures \cite{bena:brenner2009global,bena:spangenberg2016pole} or road lanes \cite{bena:hata2014}.
Nevertheless, such landmarks rarely exist in unstructured environments.

\subsection{Visual based Semantic Landmark Tracking and Mapping}
There exists a rich community in researching visual landmark tracking, especially combined with SLAM approaches.
\cite{bena:VODM} constructs point cloud models of the environment by using monocular images and inertial measurements. Afterwards, the authors compare the constructed models with known object models for the semantics.
Limitations of this approach are the need of a static environment and the processing time.
Other works have focused on semantic SLAM \cite{bena:civera2011towards,bena:nieto2010semantic}.
On the one hand, \cite{bena:zeng2018semantic} maps semantic instances at object-level by using particle-filtering for inference of the object's pose and class.
On the other hand, \cite{bena:bowman2017probabilistic} decomposes the task into estimating the data association as well as the landmark class probabilities and optimizing over the metric states.

\subsection{3D Object Detection}
Common research in object detection mainly focuses on the 2D case \cite{bena:liu2016ssd, bena:he2017mask,bena:redmon2017yolo9000}.
However, recently many 3D object detection architectures for point cloud data with or without RGB images have been developed.
Popular examples, which convert the point cloud into Bird's eye view, are the region proposal network of MV3D \cite{bena:chen2017multi} as well as Complex-YOLO \cite{bena:simon2018complex}.
Another converting example is AVOD \cite{bena:ku2017joint}, which aggregates 3D and Bird's eye view.
\cite{bena:li20173d,bena:zhou2017voxelnet} voxelize point clouds and use a volumetric CNN for 3D object detection, which is computationally demanding.
The work which is nearest to our proposed architecture is F-PointNet \cite{bena:frustum}.
It projects the point cloud into previously detected 2D objects to generate 3D-Frustums.
Afterwards, the segmentation and classification CNN PointNet and other networks are applied to generate 3D object detections.
Contrary, we propose an architecture which combines model-based methods with the strength of CNNs to account for small training data sets.

\section{Problem Description}
Our goal is to localize and classify significant landmarks in unstructured and changing environment.
Unstructured means that there are no urban structures like buildings, lane markings or signs but growing and changing vegetation.
As input we get a 3D point cloud from a depth-sensor, e.g. a LiDAR, and an image from a RGB camera.
This input can be affected by varying lighting conditions and changes in the vegetation (growing grass).
Furthermore, we only have a small training data set, as there does not exist a large data set for unstructured environments.
Therefore, we have to account for unstable detection and classification results of the CNNs.
Significant landmarks are for example static objects as trees and bushes.
Clearly, moving objects like cars are not as suitable as landmarks.
The resulting semantic landmark will be represented by a class, the Cartesian 3D position as well as the dimensions.

\section{Proposed Architecture}
This chapter explains the steps of our architecture.
Our proposed architecture consists of three main steps, namely, proposal generation, tracking and point cloud classification. The overall pipeline is shown in \Cref{fig:pipeline}.
\subsection{Proposal Generation} \label{subsec:proposal}
\begin{figure}[t!]
	\centering
	\subfigure[The output of the 2D CNN. Green 2D bounding boxes represent recognized trees, whereas red denotes a bush. Furthermore, the numbers on top of the 2D bounding boxes are the detection scores.]{\label{fig:2Ddetection}\includegraphics[width=1\linewidth]{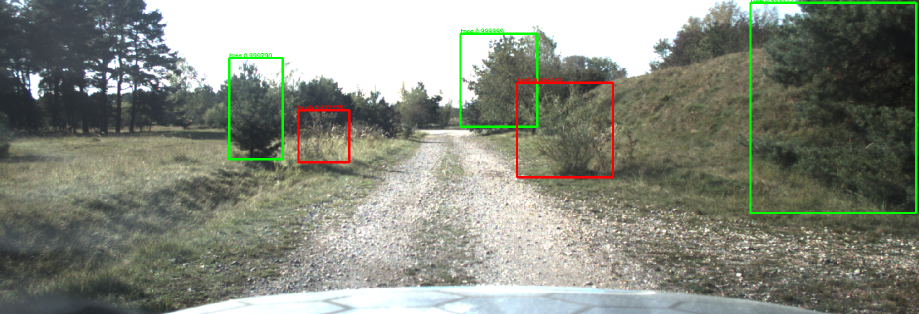}}
	\subfigure[The input point cloud as well as the segmented 3D bounding boxes of the depth sensor. Every different color denotes a corresponding object id.]{\label{fig:3Ddetection}\includegraphics[width=1\linewidth]{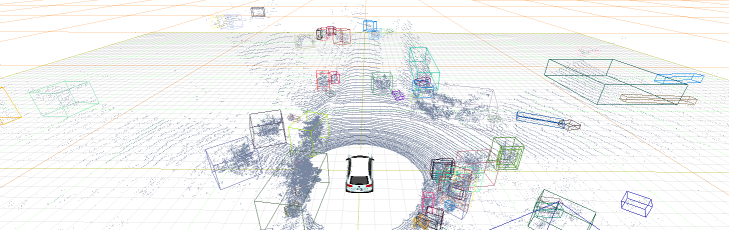}
	}
	\subfigure[An exemplary projection of 3D instances into the image plane. The green and bold rectangle denotes the classified 2D detection with its detection score at the top of it. The rectangles with slim lines are the projected 3D instances. Every instance has a different color, which only is dependent on the id. Moreover, the number on top of the rectangle displays the IoU of the instance with the 2D detection.]{\label{fig:treeproj}\includegraphics[width=\linewidth]{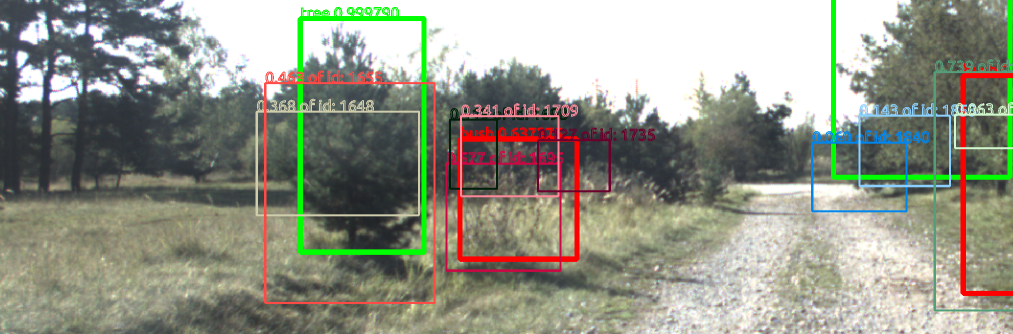}}
	\caption{The different steps of the proposal generation. Firstly, (a) shows the result of the 2D detector. Secondly, (b) visualizes the clustered point cloud. Finally, the projected 3D instances are projected into the image plane and associated with the 2D detections, as can be seen in (c).}
\end{figure}

In this step we combine 2D and 3D detections to generate classified proposals.
At first, a object detection CNN produces classified 2D detections.
We elaborated a COCO \cite{bena:COCO} pre-trained CNN, the (SSD) network \cite{bena:liu2016ssd}, and only retrained with the desired classes.
\Cref{fig:2Ddetection} shows an example output of the 2D object detector.
Simultaneously, the segmentation algorithm of \cite{bena:burger_itsc2018} clusters 3D instances from the input point cloud, which can be seen in \Cref{fig:3Ddetection}.
The instances are potentially over-segmented, see for example \Cref{fig:ddcrp1}, especially in unstructured environments.
Next, the 3D instances are projected to the image coordinate system of the RGB camera.
This is done with the extrinsic calibration between the depth sensor and camera $ H_{d}^{c} \in \mathbb{R}^{4 \times 4} $ and with the camera projection matrix $ P_{c}^{I} \in \mathbb{R}^{3 \times 4} $.
Hence, we calculate the 3D corners of the instances and project every corner to the image with:
\begin{equation}
\begin{pmatrix}\tilde{p}_{I} \\ w \end{pmatrix}
 = P_{c}^{I} \cdot H_{d}^{c} \cdot \begin{pmatrix}p_{d} \\ 1\end{pmatrix},
\end{equation}
where $ p_{d} \in \mathbb{R}^{4\times 1} $ is a corner point in LiDAR-sensor coordinates and $\tilde{p}_{I}\in \mathbb{R}^{3\times 1}$ is the corresponding non-normalized corner point in image coordinates.
Afterwards, to obtain the final point in image coordinates we normalize with:
\begin{equation}
		 p_{I} = \begin{pmatrix}
		\nicefrac{\tilde{p}_{I}^x}{w} \\ 		\nicefrac{\tilde{p}_{I}^y}{w}
		\end{pmatrix}.
\end{equation}
Next, the encasing rectangle of these corners is considered as 2D bounding box of the instance.
Then, the intersection-over-union (IoU) of the two bounding boxes is calculated to associate the 2D detections with the 3D instances.
We set the corresponding class proposal to every 3D instance with an IoU over a pre-defined threshold $ \tau \in [0,1]$.
An example can be seen in \Cref{fig:treeproj}.

\subsection{Tracking}
As shown in \Cref{fig:pipeline}, the Greedy Dirichlet Process Filter (GDPF) of \cite{bena:GDPF} is used for the tracking part.
It is a multi-target tracker with probabilistic data-association capable of handling over-segmented objects, where every component of the Dirichlet Process corresponds to a target.
A Dirichlet Process (DP) is a mixture model with an infinite number of components, however only finite ones have non-zero weight.
The DP is defined through the concentration parameter $ \alpha \in \mathbb{R} $, which controls the assignment to a new component and a random mixing measure $ G_0 $.
The GDPF consists of two main steps, namely, greedily choosing the best component for every association by the maximum conditional posterior probability for measurement to component association and updating the posterior distribution for every measurement.
\subsubsection{Basic Definitions}
Before explaining the data association and posterior update, we need some basic definitions.
For the time step $ t $, let $ \bm{Y}^i(t)= \lbrace \bm{y}_m(t)\rbrace_{m=0}^i $ be the set of previously associated measurements plus the current one.
Furthermore, $ \bm{K}_t = \lbrace k \mid k \in \mathbb{N} \rbrace $ defines the indices set of corresponding active components.
Moreover, $ j_i(t): $ $ \mathbb {N} \to \mathbb{N} $ is defined as the assignment of index $ i $ corresponding to measurement $ y_i(t) $ to the index of another measurement. Consequently, $ \bm{j}_{-i}(t) = \lbrace j_m(t) \rbrace_{m=0}^{i-1} $ defines all previous measurement-to-measurement associations.
$ z_i(t): \mathbb{N} \to \mathbb{N} $ denotes the measurement-to-component index assignment.
Furthermore, we define $ \bm{z}(t) = \lbrace z_m(t) \rbrace_{m=0}^{n_t}$, with $ n_t $ equals the number of measurements.
Every component has parameters $ \theta_{z_i(t)=k}(t)=\lbrace \bm{x}_k^d(t),\bm{P}_k(t),\epsilon_k(t)\rbrace $, with dynamical state $\bm{x}_k^d(t)$, covariance matrix $ \bm{P}_k(t) $ and existence probability $ \epsilon_k (t) $.
The dynamical state is defined as:
\begin{equation}
\bm{x}_k^{d}(t) = \begin{pmatrix}
x_k,y_k,z_k,\dot{x}_k,\dot{y}_k,\omega_k,l_k,w_k,h_k
\end{pmatrix},
\end{equation} where $ x_k,y_k,z_k $ are the Cartesian positions, $ \dot{x}_k,\dot{y}_k $ are the velocities, $ \omega $ is the yaw (heading) rate and $ l_k,w_k,h_k $ are the length, width and height of the component.
\subsubsection{Data Association}
The goal of this part is to associate the $ i $-th measurement of time step $ t $, $ \bm{y}_i(t) $, to the corresponding component $ k \in \bm{K}(t) $.
The association step utilizes the distance-dependent Chinese Restaurant Process (P) and a cluster prior.
The ddCRP is a realization of a Dirichlet Process \cite{bena:bleidistance}, where the relations between the measurements are processed.
Let $ \bm{y}_i(t),\bm{y}_m(t) $ be two measurements (in our case two 3D bounding boxes) at time step $ t $.
Then, we score their relation depending on the maximum signed distance of the bounding box center point $ y_i^{c}(t) $ and bounding box sides $ S( \bm{y}_m(t) ) = \lbrace \bm{y}_m^s(t) \rbrace_{s=0}^{5} $ of measurement $ \bm{y}_m(t) $:
\begin{equation} \label{eq:s_max}
\phi_{max}(	\bm{y}_i(t),\bm{y}_m(t)) = \max\limits_{s \in S( \bm{y}_m(t) )} \phi(y_i^{c}(t),\bm{y}_m^s(t)),
\end{equation} where the signed distance $ \phi(y_i^{c}(t),\bm{y}_m^s(t)) $ is defined with $ dist(y_i^{c}(t),\bm{y}_m^s(t)) $, the distance of $ y_i^{c}(t) $ to the side $ \bm{y}_m^s(t) \in S(\bm{y}_m(t)) $, as follows:
\begin{equation}\label{eq:s}
\phi(y_i^{c}(t),\bm{y}_m^s(t)) = \begin{cases}
dist(y_i^{c}(t),\bm{y}_m^s(t)) & \text{if $ \bm{y}_m(t) $ contains $ y_i^{c}(t) $} \\
-dist(y_i^{c}(t),\bm{y}_m^s(t)) & \text{else}.
\end{cases}
\end{equation}
With the definition of \Cref{eq:s_max,eq:s}, the following scoring function $ d_{im}(\bm{y}_i (t),\bm{y}_m (t) ) $ is used:
\begin{equation} \label{eq:sdist}
d_{im}(\bm{y}_i (t),\bm{y}_m (t) ) = \frac{1}{ 1 + e^{ -0.75( \phi_{max}(	\bm{y}_i(t),\bm{y}_m(t)) + 1 )  }}.
\end{equation}
\begin{figure}[t!]
	\begin{center}
		\newcommand{\cuboid}[9]{
    \draw[#7] (#1, #2, #3) -- ++(#4, 0, 0) -- ++(0, 0, #6) -- ++(-#4, 0, 0) -- cycle;
    \draw[#8] (#1, #2, #3) -- ++(#4, 0, 0) -- ++(0, #5, 0) -- ++(-#4, 0, 0) -- cycle;
    \draw[#8] (#1, #2, #3) -- ++(0, 0, #6) -- ++(0, #5, 0) -- ++(0, 0, -#6) -- cycle;
    \draw[#9] (#1+#4, #2, #3) -- ++(0, 0, #6) -- ++(0, #5, 0) -- ++(0, 0, -#6) -- cycle;
    \draw[#8] (#1, #2, #3+#6) -- ++(#4, 0, 0) -- ++(0, #5, 0) -- ++(-#4, 0, 0) -- cycle;
    \draw[#8] (#1, #2+#5, #3) -- ++(#4, 0, 0) -- ++(0, 0, #6) -- ++(-#4, 0, 0) -- cycle;
}

\begin{tikzpicture}
\tikzstyle{bottomfill} = [thick, fill=lightgray, fill opacity=.8]
\tikzstyle{bottomfill1} = [thick, fill=lightgray, fill opacity=.6]
\tikzstyle{bottomfill2} = [thick, fill=blue, fill opacity=.4]
\tikzstyle{camerafill} = [thick, fill=red!20, fill opacity=.4]
\tikzstyle{camerafill1} = [thick, fill=green!20, fill opacity=.4]

\cuboid{0}{0}{0}{3}{2.5}{2}{bottomfill}{bottomfill1}{bottomfill2}
\cuboid{1.2}{0.9}{0.45}{1.0}{0.8}{1.2}{camerafill}{camerafill}{camerafill}
\cuboid{4}{0.5}{2}{1.5}{0.6}{1}{camerafill1}{camerafill1}{camerafill1}

\node at (3.5,1.7,2.1) {\small $ \bm{y}_m^s $};
\node at (-0.1,2.7,0) {\small $ \bm{y}_m $};
\node at (0.8,1.6,0.9) {\small $ \bm{y}_1 $};
\node at (4.2,1.3,2) {\small $ \bm{y}_2 $};

\draw[fill=black] (1.9, 1.45, 1.65) circle (0.1em) node[below] {\small $ y_1^{c} $};
\draw[->,dotted,thick] (1.9, 1.45, 1.65)  -- ++(1.2, 0.00, 0.0);
\draw[fill=black] (4.5, 0.6, 2) circle (0.1em) node[below] {\small $ y_2^{c} $};
\draw[->,dotted,thick] (4.5, 0.6, 2.)  -- ++(-1.2, 0.00, 0.0);
\end{tikzpicture}
		\caption{Possible relations between different 3D bounding boxes. The gray bounding box indicates a previously associated measurement $ \bm{y}_m $, whereas the red $ \bm{y}_1 $ and green $ \bm{y}_2 $ bounding boxes are candidates for association. The dotted arrows indicate the maximum signed distance of center points $ \bm{y}_1^c $ and $ \bm{y}_2^c $  (\Cref{eq:s_max}) to the corresponding bounding box side $ \bm{y}_m^s $(blue). On the one hand, $ \bm{y}_1^c $ lies inside $ \bm{y}_m $, therefore, the signed distance will be positive. On the other hand, $ \bm{y}_m $ does not contain $ \bm{y}_2^c $. Hence, the corresponding signed distance will be negative.} \label{fig:meas_rel}
		\vspace{-0.5cm}
	\end{center}
\end{figure}
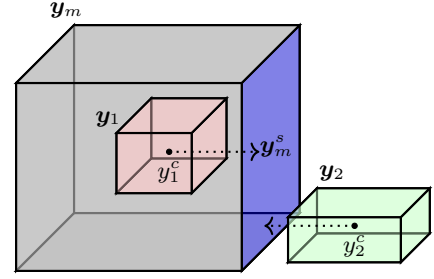
Measurement relations are more likely if the distance is less negative or even positive.
\Cref{fig:meas_rel} illustrates two exemplary cases of measurement relations.
Finally, we calculate the ddCRP as follows \cite{bena:GDPF}:
\begin{equation}
\Ptwo(j_i(t) = m|\bm{j}_{-i},\alpha) \propto \begin{cases}
d_{im}(\bm{y}_i (t),\bm{y}_m (t) )  &i \neq m, \\
\alpha  &i=m
\end{cases}. \label{eq:ddcrp}
\end{equation}
\begin{figure}[t!]
	\centering
		\subfigure[Different 3D instances are associated to the Dirichlet components according to the ddCRP. Even though the bush on the right side is heavily oversegmented, the corresponding instances (red bounding boxes on the right side) are associated to the component which represents the bush.]{\label{fig:ddcrp1}
			\includegraphics[width=1\linewidth]{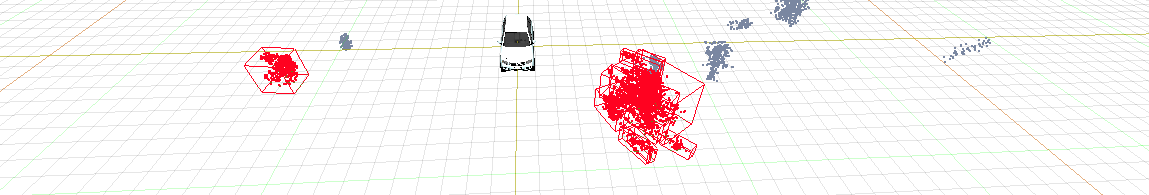}}
		\subfigure[The cluster prior visualized as ellipsoid. It is dependent on the current component's dimension plus their corresponding uncertainty resulting from the last time step.]{\label{fig:prior1}
			\includegraphics[width=1\linewidth]{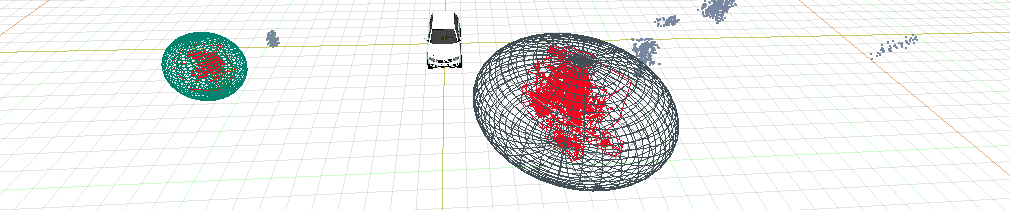}}
		\caption{The data association step result of the GDPF. 3D bounding boxes originate from the clustering algorithm. The colors of the ellipses correspond to the components id, whereas the color of the bounding boxes indicates classes.}
		\vspace{-0.2cm}
\end{figure}
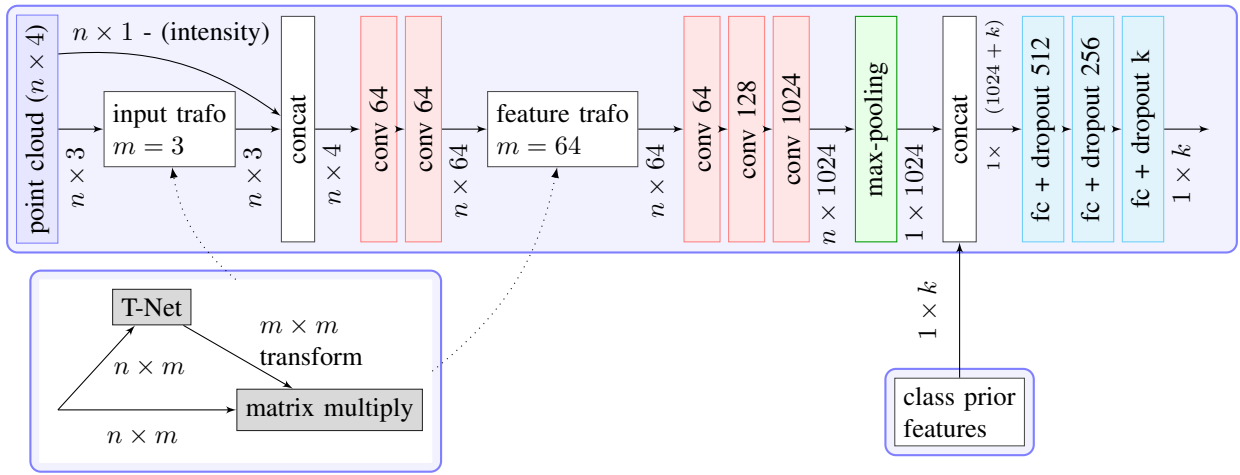
\begin{figure*}[t!]
	\begin{center}
		\newsavebox\inputtrafo
\begin{lrbox}{\inputtrafo}
	\begin{tikzpicture}[]
		\tikzstyle{box}=[shape=rectangle,draw=green!60!black,fill=green!10!black,text width=1em]
		\tikzstyle{whitebox}=[shape=rectangle,draw=white,fill=white]
		\tikzset{
			box/.style={rectangle, minimum width =60pt, minimum height=40pt,text width=5em,text centered, draw=gray!30!black,fill=gray!30!white,node distance = 10mm}
		}
		\node[whitebox](center_input_t){};
	    \node[whitebox](midway)[right=of center_input_t]{};
		\node[rectangle,draw=gray!30!black,fill=gray!30!white](T-Net)[above=of midway]{T-Net};
		\node[rectangle,draw=gray!30!black,fill=gray!30!white](MM)[right=of midway]{matrix multiply};

		\path [line] (center_input_t.center) -- node [midway,below=0.3em,align=left] {$ n \times m $} (MM);
		\path [line] (center_input_t.center) -- node [midway,right=0.3em,align=left] {$ n \times m $} (T-Net);
		\path [line] (T-Net) -- node [near start,right=1.5em,align=left] {$ m \times m $\\ transform} (MM);
	\end{tikzpicture}
\end{lrbox}

%

\begin{tikzpicture}
\tikzstyle{bigbox} = [draw=blue!50, thick, fill=blue!5, rounded corners, rectangle]
\tikzstyle{input}=[shape=rectangle,draw=blue!50,fill=blue!10,minimum height=3cm]
\tikzstyle{conv2d}=[shape=rectangle,draw=red!50,fill=red!10,minimum height=3cm]
\tikzstyle{feature}=[shape=rectangle,draw=white!30!black,fill=white!30!white]
\tikzstyle{bluebox}=[shape=rectangle,draw=blue!5,fill=blue!5]
\tikzstyle{maxp}=[shape=rectangle,draw=green!60!black,fill=green!10,minimum height=3cm]
\tikzstyle{trafo}=[<-,dotted]
\tikzstyle{fc}=[shape=rectangle,draw=cyan!50,fill=cyan!10,minimum height=3cm]
\tikzstyle{whitebox}=[shape=rectangle,draw=white,fill=white]

    \node[input] (input){\rotatebox{90}{point cloud ($ n \times 4 $)}};
    \node[shape=rectangle,draw=white!30!black,fill=white!30!white,minimum height=1cm,align=left](input_trafo)[right=0.6cm of input]{input trafo \\ $ m=3 $};

    \node[whitebox](t-net)[below=2cm of input_trafo.east]{\usebox\inputtrafo};        \path [line] (input) --node [midway,below=0.3em,align=left] {\rotatebox{90}{$ n \times 3 $}}(input_trafo);
    \draw [arrow,dotted,bend left=20] (t-net.north) to (input_trafo);
    \node[feature,minimum height=3cm] (input_feature)[right=0.6cm of input_trafo]{\rotatebox{90}{concat}};
   \draw [line,bend left=20] (input.east)+(0,+1) to node [midway,above=0.3em,align=right] {$ n \times 1 $ - (intensity)} (input_feature);

    \path [line] (input_trafo) --node [midway,below=0.3em,align=left] {\rotatebox{90}{$ n \times 3 $}} (input_feature);
    \node[conv2d](conv2d64)[right=0.6cm of input_feature]{\rotatebox{90}{conv 64}};
    \path [line] (input_feature) -- node [midway,below=0.2em,align=left] {\rotatebox{90}{$ n \times 4 $}}(conv2d64);
    \node[conv2d](conv2d642)[right=0.1cm of conv2d64]{\rotatebox{90}{conv 64}};
    \path [line] (conv2d64) -- (conv2d642);
     \node[shape=rectangle,draw=white!30!black,fill=white!30!white,minimum height=1cm,align=left](feature_trafo)[right=0.6cm of conv2d642]{feature trafo \\ $ m=64 $};
    \path [line] (conv2d642) --node [midway,below=0.2em,align=left] {\rotatebox{90}{$ n \times 64 $}}(feature_trafo);
    \draw [arrow,dotted,bend right=20] (t-net.east) to (feature_trafo);
    \node[conv2d](conv2d643)[right=0.6cm of feature_trafo]{\rotatebox{90}{conv 64}};
    \path [line] (feature_trafo) --node [midway,below=0.2em,align=left] {\rotatebox{90}{$ n \times 64 $}} (conv2d643);
    \node[conv2d](conv2d128)[right=0.1cm of conv2d643]{\rotatebox{90}{conv 128}};
    \path [line] (conv2d643) -- (conv2d128);
    \node[conv2d](conv2d1024)[right=0.1cm of conv2d128]{\rotatebox{90}{conv 1024}};
    \path [line] (conv2d128) -- (conv2d1024);
    \node[maxp](maxpool)[right=0.6cm of conv2d1024]{\rotatebox{90}{max-pooling}};
    \path [line] (conv2d1024) --node [midway,below=0.2em,align=left] {\rotatebox{90}{$ n\times 1024 $}} (maxpool);
    \node[feature,minimum height=3cm] (conc_feature)[right=0.6cm of maxpool]{\rotatebox{90}{concat}};
    \node[feature,align=left] (onehot)[below=1.8cm of conc_feature]{class prior \\ features};
    \path [line] (maxpool) --node [midway,below=0.2em,align=left] {\rotatebox{90}{$ 1\times1024 $}} (conc_feature);
    \path [line] (onehot) --node [midway,left=0.1em,align=left] {\rotatebox{90}{$1\times k $}} (conc_feature);
    \node[fc](fc512)[right=0.6cm of conc_feature]{\rotatebox{90}{fc + dropout 512}};
    \path [line] (conc_feature) --node [midway,below=0.2em,align=left] {\rotatebox{90}{\scriptsize$1\times $}} node [midway,above=0.2em,align=left] {\rotatebox{90}{\scriptsize$(1024+k) $}} (fc512);
    \node[fc](fc256)[right=0.1cm of fc512]{\rotatebox{90}{fc + dropout 256}};
    \path [line] (fc512) -- (fc256);
    \node[fc](fck)[right=0.1cm of fc256]{\rotatebox{90}{fc + dropout k}};
    \path [line] (fc256) -- (fck);
    \node[bluebox](result)[right=0.6cm of fck]{};
    \path [line] (fck) --node [midway,below=0.2em,align=left] {\rotatebox{90}{$1\times k$}} (result);
    \begin{pgfonlayer}{background}
        \node[bigbox] [fit = (input) (result)] {};
    \end{pgfonlayer}
    \begin{pgfonlayer}{background}
	    \node[bigbox] [fit = (t-net)] {};
    \end{pgfonlayer}
    \begin{pgfonlayer}{background}
        \node[bigbox] [fit = (onehot)] {};
    \end{pgfonlayer}
\end{tikzpicture}
		\caption{The proposed network is a modified PointNet \cite{bena:pointnet}. We added the intensity as 4th input layer and the class prior of the 2D object detection or the last time step. The T-Net \cite{bena:pointnet} learns a transformation matrix for robust estimation results, which is applied to every point.} \label{fig:modpointnet}
	\end{center}
\end{figure*}
The cluster prior is modeled as an ellipse at the x-y plane, depending on the component $ k $'s dynamical state $\bm{x}_k^{d}(t)$ and covariance.
Define $ a_k= l_k + \sqrt{\bm{P}_k^{l l}} $ and $ b_k= w_k + \sqrt{\bm{P}_k^{w w}} $ with state covariance elements $ \bm{P}_k^{ll},\bm{P}_k^{ww} $ of length $ l_k $, respectively width $ w_k $.
Then, for measurement $ \bm{y}_i(t) $ with its Cartesian x and y-positions $ y_i^x(t),y_i^y(t) $, the cluster prior is calculated through \cite{bena:GDPF}:
\begin{equation} \label{eq:car_prior}
\pi_{z_i(t)=k} =	e^{\left( -\left( \frac{(x_k - \bm{y}_i^x (t))^2}{a_k}+\frac{(y_k - \bm{y}_i^y (t))^2}{b_k}  \right)\right)}.
\end{equation}

Let $ m_k \in \mathbb{N} $ be the measurement index associated to the component $ k \in \bm{K}_t $ and $ \tilde{\bm{K}}_t = \bm{K}_t \cup \lbrace k_{new}\rbrace $ with $ k_{new} \in \mathbb{N} $ be a newly created component index.
Then, the measurement-to-component association is calculated through the following conditional posterior probability \cite{bena:GDPF}:
\begin{multline}
\Ptree(z_i(t) = k|\bm{Y}^{i}(t),\bm{j}_{-i},\alpha) \propto \\
	\frac{\Ptwo(j_i = m_k|\bm{j}_{-i},\alpha ) \cdot \pi_{z_i(t)=k} }{\sum_{n\in \tilde{\bm{K}}_t}\Ptwo(j_i = m_n|\bm{j}_{-i},\alpha ) \cdot \pi_{z_i(t)=n}}. \label{eq:label}
\end{multline}
\subsubsection{Update the Posterior Distribution}
After assigning the measurement to the best matching component, the posterior distribution has to be updated.
The posterior distribution is defined as \cite{bena:GDPF}:
\begin{align}
\Ptwot[\bm{\theta}_{z_i(t)}(t)|\bm{y}_{i-1}(t) ] &\propto G_0(\bm{\theta}_{z_i(t)=k}(t)) \nonumber\\
&\cdot \Ptwot[ \bm{y}_i(t)|\bm{\theta}_{z_i(t)=k}(t)  ] \\
&\cdot \Ptwot[ \bm{\theta}_{z_i(t)=k}(t )|\bm{\theta}_{z_i(t-1)=k}(t-1 )], \nonumber
\end{align}
where the first part generates new components with the base prior distribution $ G_0 $, the second part involves the dynamic parameters and other component parameters, \eg existence probability. Finally, the third part models the time evolution of the component parameters.
For the dynamical part of the component the filter utilizes Kalman Filtering.
Therefore, we have to choose a process and measurement model.
We chose the Augmented Coordinated Turn \cite{bena:BarShalom02} (ACT) as process model as it can appropriately handle landmarks, persons and cars.
As measurement models we have the simple 3D position and dimension measurements.
\subsection{Point Cloud Classification}
\begin{figure}[t!]
\centering
		\subfigure[Original points of the section.]{\includegraphics[width=\linewidth]{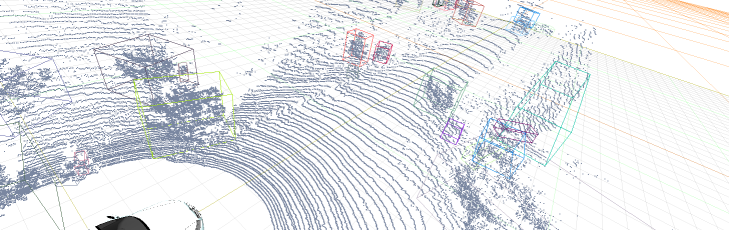}}
		\subfigure[Randomly sampled 1024 points per component.]{
				\includegraphics[width=\linewidth]{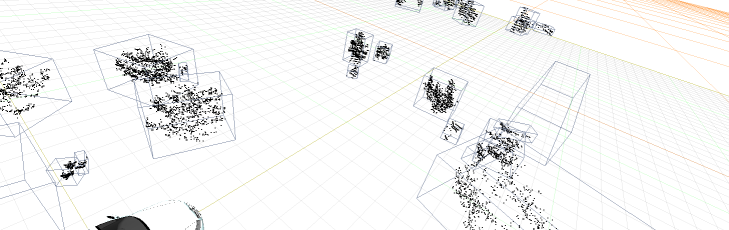}}
		\subfigure[Randomly sampled 512 points per component.]{\includegraphics[width=\linewidth]{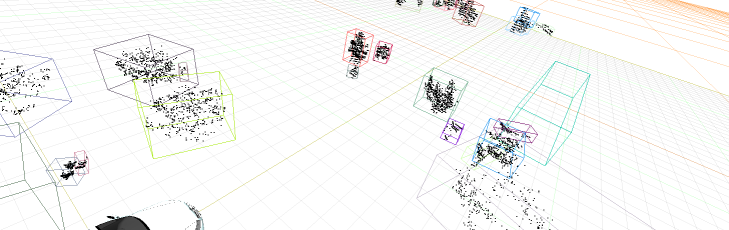}}
		\caption{Visualizing sampled points with varying numbers. (a) shows the original and dense point cloud with segmented bounding boxes, whereas (b) and (c) visualize the randomly samples points of the corresponding components.} \label{fig:comp_samples}
\end{figure}
In this part we classify the accumulated point cloud for each component with a class prior.
The accumulated point cloud's size varies between 25 and 12000 points.
Therefore, we have to randomly sample a fixed number of points $ n_p \in \mathbb{N}$, as our classifier needs a fixed $ n_p $ as input layer.
\Cref{fig:comp_samples} shows the effects with sampling 512 and 1024 points per component.
The class prior could be the classification result from the last time step or the class proposal of the proposal generation step, explained in \Cref{subsec:proposal}.
The base of the point cloud classification is the classifying PointNet of \cite{bena:pointnet}.
Similar to the segmentation network in \cite{bena:frustum} we integrate the intensity as 4th input layer and concatenate the class prior scores to the features after the max-pooling layer.
\Cref{fig:modpointnet} shows the modified PointNet structure.


\section{Results} \label{sec:results}
\begin{figure}[t!]
	\centering
	\subfigure[Inferred boxes (red) and ground truth (blue) on trained data. ]{\label{fig:yolo_train}\includegraphics[width=0.98\linewidth]{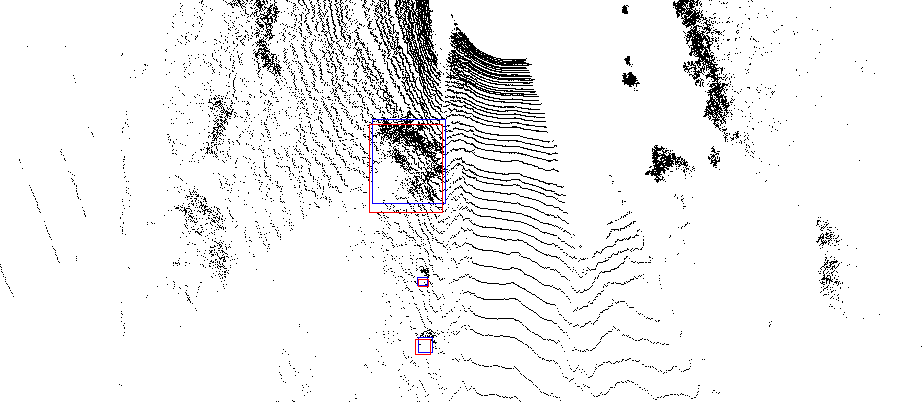}}
	\subfigure[Inferred boxes (red) and ground truth (blue) during the testing phase (unseen data). Complex-YOLO is not capable of predicting correct boxes on unseen point clouds (snippet of the output). ]{\label{fig:yolo_det}\includegraphics[width=0.98\linewidth]{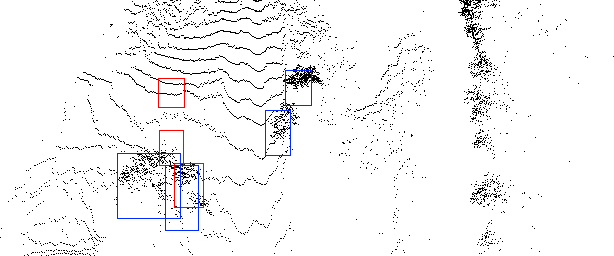}
	}
	\caption{The capability of Complex-YOLO \cite{bena:simon2018complex} with a small training data set. In (a) the predicted boxes are near the ground-truth boxes. Nevertheless, the network could not generalize enough with a small training data set to predict correct boxes for unseen data as can be seen in (b).} \label{fig:complex_yolo}
\end{figure}
\begin{figure}[ht!]
	\centering
	\subfigure[The output of the 2D object detector with the classes bush (red) and tree (green).]{\label{fig:2d_det}\includegraphics[width=1\linewidth]{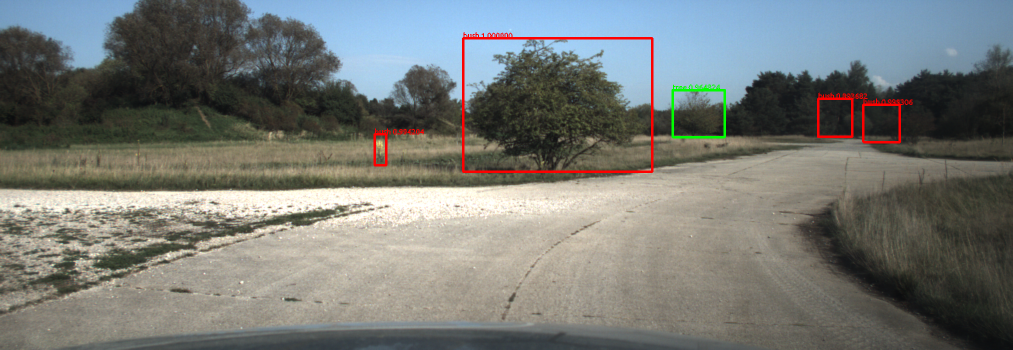}}
	\subfigure[The result of our approach (red bounding box) compared against the ground truth (blue bounding box)]{\label{fig:our}\includegraphics[ width=1\linewidth]{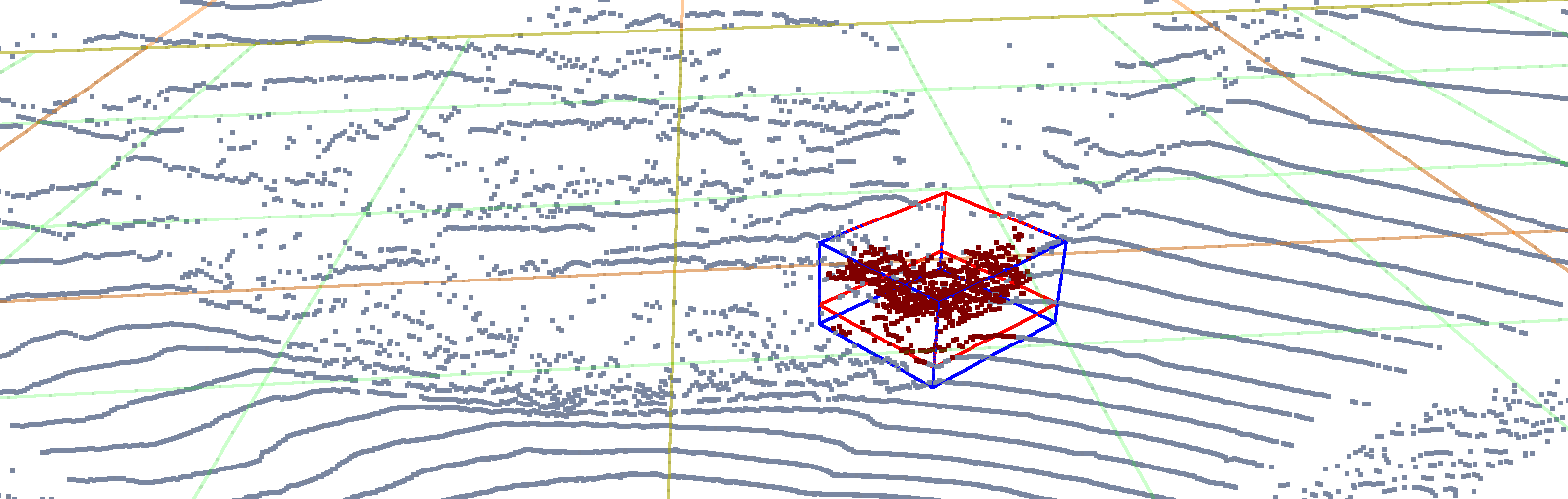}}
	\subfigure[3D detection results of F-PointNet. There are many false positives (bounding boxes without colored points) and the nearest bounding box to the ground truth has a large position error. The large position error could be caused by the occlusion handling capability of F-PointNet. ]{\label{fig:point1024}\includegraphics[width=1\linewidth]{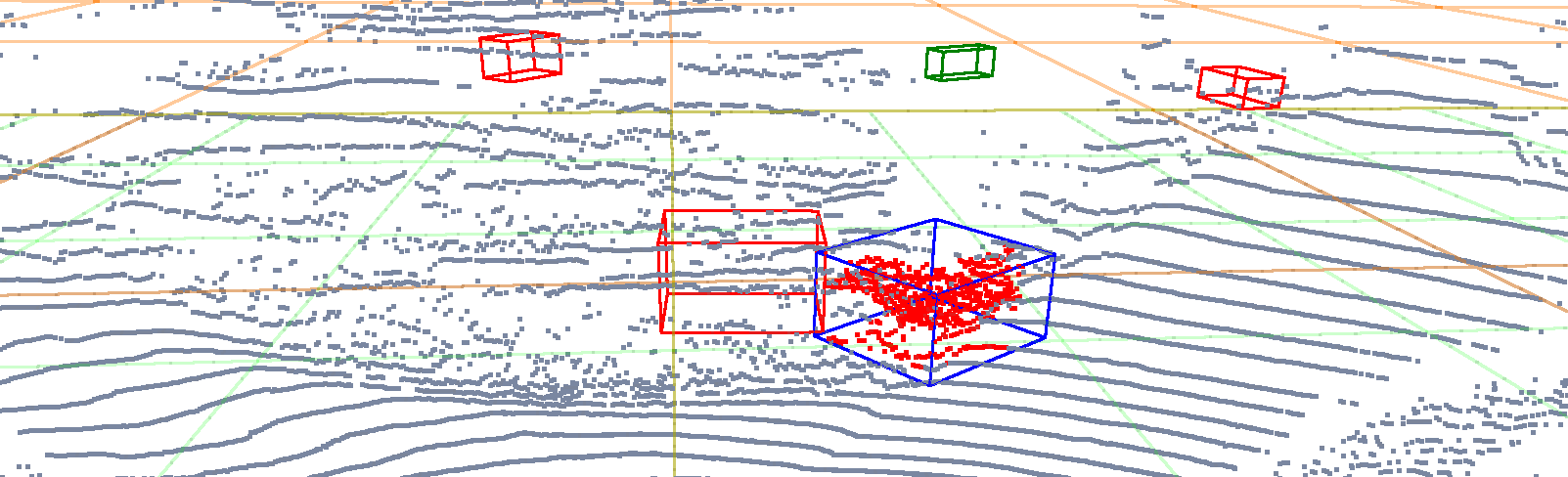}
	}
	\caption{Qualitative results of our approach and F-PointNet. The blue bounding boxes are the ground truth, whereas red and green bounding boxes are results of the methods.} \label{fig:comp_our_point}
\end{figure}
\begin{figure}[ht!]
	\centering
	\subfigure[The output of the 2D object detector with the classes bush (red) and tree (green).]{\label{fig:2d_det_wrong}\includegraphics[width=1\linewidth]{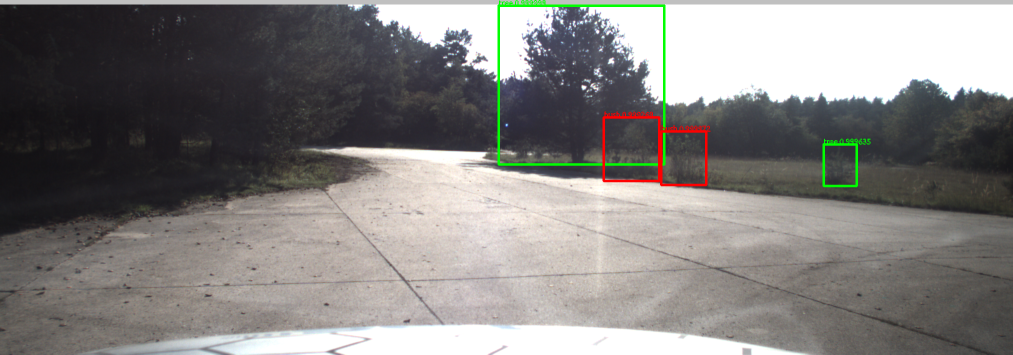}}
\subfigure[Result of our method compared to ground truth. There is one false negative (blue bounding box with black circle) as the landmark has never been detected in the 2D case.]{\label{fig:point1024_wrong}\includegraphics[width=1\linewidth]{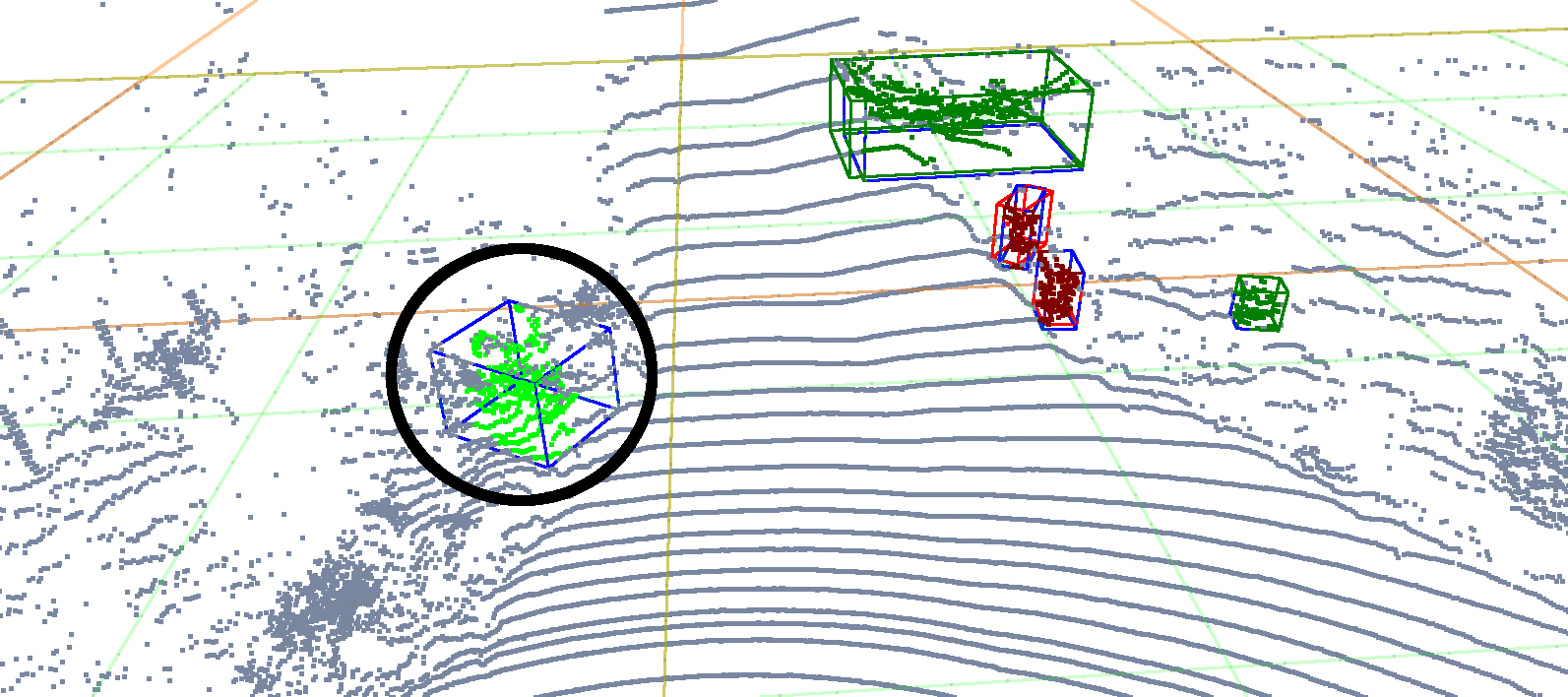}
	}
	\caption{An example of our method with a false negative.} \label{fig:comp_our_fales}
\end{figure}
The results were made in a real-world scenario in unstructured environments.
Data has been recorded with a roof-mounted Velodyne HDL-64 S2 \cite{velodyne-lidar}, which approximately records 130k points per time step, and a RGB camera behind the windshield.
Our goal is to detect 3D bounding boxes of significant landmarks (trees and bushes) with their Cartesian positions, dimension and their type.
Video footage can be found online\footnote{https://youtu.be/RwoRqmYzpzs}.
Furthermore, there only exist $ 419 $ frames of data, manually labeled at our institute, which we split in $ 80\% $ training, $ 10\% $ validation and $ 10\% $ testing data.
The evaluation was done on a standard office computer with an Intel(R) Core i7 CPU and the consumer graphic card NVIDIA GeForce GTX 1060.
For evaluation, the metrics precision $ \mathcal{P} $ and recall $ \mathcal{R} $ are used.
They are calculated with true positives $ t_p $, false positives $ f_p $ and false negatives $ f_n $ as follows:
\begin{align}
\mathcal{P} &= \frac{t_p}{f_n + f_p},\\
\mathcal{R} &= \frac{t_p}{t_p + f_p}.
\end{align}
Furthermore, the root mean squared error (RMSE), with respect to the Cartesian $ x $- and $ y $-positions in meter, is compared.
We calculate the RMSE relative to the hand-labeled ground-truth bounding-boxes constructed around the point cloud data of the object.
\Cref{fig:comp_our_point} visualizes the ground-truth as blue bounding-box.
The z-position is neglected as VoxelNet only generates results in the $ x-y $ plane.
We define a detection as true positive if it has the same class as the ground-truth, and the ground-truth's bounding box contains the bounding box center of the detection.
Furthermore, we compare the 3D bounding box overlap of the ground-truth and the corresponding detection.

Our evaluation is divided into three parts.
Firstly, we quantitatively and qualitatively compare with state-of-the-art methods for 3D object detection, which, i.\,e.,\ have proven to generate excellent results at the KITTI benchmark \cite{bena:KITTI}.
Secondly, we evaluate our approach with different parameter choices.
Finally, the run-time of our approach is compared with the other methods.
\subsection{Performance Comparison}
We test our approach against the popular methods Complex-YOLO \cite{bena:simon2018complex} and Voxelnet \cite{bena:zhou2017voxelnet}, which purely rely on 3D point clouds as well as F-PointNet \cite{bena:frustum}. F-PointNet also generates proposals through a 2D object detector.
Therefore, we used the same 2D detections as in our method.
It can be seen in \Cref{tab:comparison_all} that our method generates significantly better results especially against the pure point cloud based methods.
Hence, we can conclude that both are not suitable for scenarios with a small training set.
An exemplary output of Complex-YOLO is shown in \Cref{fig:complex_yolo}.
The closest method to ours is F-PointNet (v1), albeit we even outperform this method by large margins.
Moreover, in \Cref{fig:comp_our_point} we visualize results of one time step of our method and F-PointNet.
F-PointNet has more false positives than our method as well as less bounding box overlap.
Summarizing, our method shows competitive results against current state-of-the-art algorithms.
Nevertheless, we observe false negatives, if the 2D detector never detects the landmark.
This can be seen in the failure case of \Cref{fig:comp_our_fales}.
\subsection{Parameter Analysis}
We evaluate our results with respect to the number of points per component $ n_p $ and the IoU threshold $ \tau $ in the proposal generation.
\Cref{tab:compare_param} shows that the overall best results are taken with $ \tau = 0.2 $ and $ n_p=1024 $.

\section{Runtime Comparison}
\Cref{tab:runtime} shows the run-time evaluation of the different methods.
Our method shows real-time capability. In the case of $ n_p = 512 $, the mean run-time is about 72ms. On average, it consists of 25ms point cloud clustering, 23ms proposal generation, 13ms tracking and 11ms point cloud classification.
Complex-YOLO is the fastest algorithm, however, it generates worse results and only works on a cropped area of the point cloud.
\begin{table}
	\vspace{0.1cm}
	\centering
	\caption{Run-time comparison of the evaluated methods. The fastest method is Complex-YOLO but it operates on a cropped view of the point cloud (40mx40m). Contrary, VoxelNet is the slowest method and definitely not real-time capable. F-PointNet (v1) operates only on the limited view of their region proposals. Nevertheless, our methods are faster and real-time capable. Moreover, we operate on the whole point cloud with approximately 130k points.}\label{tab:runtime}
	\begin{tabular}{l|r}
		Method & $ \varnothing t $ in ms \\
		\hline
		Ours ($ n_p= $512) & 72\\
		Ours ($ n_p= $1024) & 92\\
		VoxelNet & 2890\\
		Complex-YOLO (cropped) & 39\\
		F-PointNet (v1) & 132
	\end{tabular}
\end{table}

\begin{table*}[ht!]
		\vspace{0.1cm}
		\centering
		\caption{Comparison of our approaches with renowned methods. Complex-YOLO \cite{bena:simon2018complex} and Voxelnet \cite{bena:zhou2017voxelnet} are pure point cloud based methods, whereas F-PointNet \cite{bena:frustum} utilizes images and point clouds. Our method outperforms the pure learning-based methods in all categories.} \label{tab:comparison_all}
		\begin{tabular}{|l|cccc|cccc|}
			\hline
			\multirow{2}{*}{Method} & \multicolumn{4}{c|}{Trees} & \multicolumn{4}{c|}{Bushes}\\
			& $ \mathcal{R} $ & $ \mathcal{P} $ & $ RMSE$  & $ \varnothing$overlap  & $ \mathcal{R} $ & $ \mathcal{P} $ & $ RMSE$ & $ \varnothing$overlap  \\
			\hline
			Complex-YOLO \cite{bena:simon2018complex}& $ 0.01 $&0.5&1.02&-&0.06&0.44&1.02&- \\
			Voxelnet \cite{bena:zhou2017voxelnet}& $ 0 $&$ 0 $&-&-&$ 0 $&$ 0 $&-&- \\
			F-PointNet(v1 512) \cite{bena:frustum}  & $ 0.51 $&$ 0.87 $&$ 1.05 $&$ 0.24 $&$ 0.39 $&$ 0.55 $&$ 1.29 $&$ 0.21 $ \\
			F-PointNet(v1 1024) \cite{bena:frustum} & $ 0.58 $&$ 0.88 $&$ 0.90 $&$ 0.25 $&$ \bm{0.57} $&$ 0.64 $&$ 1.01 $&$ 0.25 $ \\
			F-PointNet(v2 512) \cite{bena:frustum} & $ 0 $&$ 0 $&-&-&$ 0 $&$ 0 $&-&- \\
			F-PointNet(v2 1024) \cite{bena:frustum} & $ 0 $&$ 0 $&-&-&$ 0 $&$ 0 $&-&- \\
			Ours (best 512)  &$ \bm{0.78} $&$ \bm{0.89} $&$ \bm{0.58} $&$ \bm{0.68} $&$ \bm{0.57} $&$ \bm{0.94} $&$ \bm{0.04} $&$ \bm{0.92} $ \\
			Ours (best 1024)  &$ \bm{0.78}$ &$ \bm{0.89} $&$ \bm{0.58} $&$ \bm{0.68} $&$ \bm{0.57} $&$ \bm{0.94} $&$ \bm{0.04} $&$ \bm{0.93} $ \\
			\hline
		\end{tabular}
\end{table*}
\begin{table*}[ht!]
\centering
		\caption{Comparing our results with different IoU-threshold $ \tau $ and number of points $ n_p $. Clearly, the precision grows with higher $ \tau $ as there are less proposals generated. Contrary, the recall significantly grows with smaller $ \tau $. Altogether, $ n_p = 1024 $ generates better results but only with a small margin.}\label{tab:compare_param}
		\begin{tabular}{|lc|cccc|cccc|}
			\hline
			\multirow{ 2}{*}{$\tau$} &\multirow{ 2}{*}{$ n_p $}& \multicolumn{4}{c|}{Trees} & \multicolumn{4}{c|}{Bushes}\\
			&& $ \mathcal{R} $ & $ \mathcal{P} $ & $ RMSE$  & $ \varnothing$overlap  & $ \mathcal{R} $ & $ \mathcal{P} $ & $ RMSE$ & $ \varnothing$overlap \\
			\hline
			$ 0.2 $ &$ 512 $ &$ \bm{0.78} $&$ 0.89 $&$ \bm{0.58} $&$ \bm{0.68} $&$ 0.57 $&$ 0.94 $&$ \bm{0.04} $&$ 0.92 $ \\
			$ 0.3 $ &$ 512 $ &$ 0.67 $&$ 0.87 $&$ 0.64 $&$ 0.67 $&$ 0.64 $&$ \bm{0.95} $&$ 0.09 $&$ 0.89 $ \\
			$ 0.4 $ &$ 512 $ &$ 0.56$ &$0.95$  &$0.79$  &$0.46  $&$ 0.61 $&$ 0.94 $&$ 0.24 $&$ 0.68 $ \\
			$ 0.2 $ &$ 1024 $&$ \bm{0.78}$ &$ 0.89 $&$ \bm{0.58} $&$ \bm{0.68} $&$ 0.57 $&$ 0.94 $&$  \bm{0.04} $&$ \bm{0.93} $ \\
			$ 0.3 $ &$ 1024 $& $0.66$ &$ 0.87 $&$ \bm{0.58} $&$ 0.58 $&$ \bm{0.68} $&$ \bm{0.95} $&$ 0.09 $&$ 0.88 $ \\
			$ 0.4 $ &$ 1024 $& $0.56$ &$\bm{0.96}$  &$0.78$  &$0.60$  &$ 0.61$ &$ 0.94 $&$ 0.15 $&$ 0.88 $ \\
			\hline
		\end{tabular}
\end{table*}

\section{Conclusion} \label{sec:conclusion}
In this paper, we propose a semantic landmark detection architecture.
It is capable to robustly detect and track landmarks in unstructured environments.
By enhancing Deep Learning based classifiers with model-based segmentation algorithms and recursive filtering, it showed robustness against fluctuating detection results, which occur, \eg due to small training data-sets.
We showed superior performance against state-of-the-art 3D object detection algorithms.
Furthermore, our approach is real-time capable even on a consumer graphic card like the NVIDIA GeForce GTX 1060.

Future work should focus on the extension of detected landmark types.
Moreover, a class specific cluster prior could further enhance the tracking performance.

\section*{ACKNOWLEDGMENT}
The authors gratefully acknowledge funding by the Federal Office of Bundeswehr Equipment, Information Technology and In-Service Support (BAAINBw).

\bibliographystyle{IEEEtran}
\bibliography{additional_abrv,bena}
\end{document}